# QF: Quick Feedforward AI Model Training without Gradient Back Propagation

**Feng Qi**
OxInnovate
1206604664@qq.com

## Abstract

We propose Quick Feedforward (QF) Learning, a novel knowledge consolidation framework for transformer-based models that enables efficient transfer of instruction-derived knowledge into model weights through feedforward activations—without any gradient backpropagation. Unlike traditional fine-tuning, QF updates are computed in closed form, require minimal parameter modification, and preserve prior knowledge. Importantly, QF allows models to train and infer within the same runtime environment, making the process more resource-efficient and closely aligned with how the human brain operates. Code and models are open sourced on GitHub. I hope QF Learning inspires a more efficient and brain-like paradigm for AI systems.

## 1    Introduction

Gradient backpropagation has been the cornerstone algorithm driving the success of artificial intelligence (AI) models. However, despite its wide adoption, gradient backpropagation faces fundamental limitations. Firstly, there is little neuroscientific evidence suggesting that human brains employ gradient backpropagation mechanisms, particularly over long-range neural connections. It remains biologically implausible that humans adjust synaptic weights through cascades of precise gradient calculations. Moreover, backpropagation typically requires up to four times the memory overhead to store auxiliary variables such as momentum and gradients—an architectural demand not observed in the human brain. Unlike artificial models, the brain does not separate training from inference; it learns and reasons in an integrated, continuous manner. Also, human learning typically occurs incrementally, acquiring knowledge piece-by-piece without significant forgetting. In contrast, gradient-based methods often require extensive fine-tuning on large datasets to integrate even one new fact.

Recent research has attempted to overcome these limitations through knowledge editing methods that directly modify specific factual representations within AI models. Prominent approaches include ROME [1], which edits model parameters via targeted updates, and knowledge neurons [2], which identify and manipulate neurons that encode specific factual information. Yet, these approaches universally rely on gradient backpropagation at critical steps: ROME's method calculates value vector $v^*$ via gradient descent, Knowledge Neurons attribute scoring and neuron identification rely on gradient computations, and knowledge editing [3] hypernetworks also require gradients to optimize their parameters $\varphi$.

In this paper, we propose QF learning, a novel paradigm that bypasses gradient backpropagation by using closed-form feedforward updates to inject new knowledge into model weights. QF enables training and inference within the same runtime environment, avoiding model reloading and reducing extra computational cost for faster, more efficient learning. A single example suffices to learn a new fact, without large datasets or iterative tuning. Because QF updates involve minimal changes to model weights, it effectively avoids catastrophic forgetting. It thus offers a biologically inspired, efficient, and flexible alternative for updating and inferring AI models.

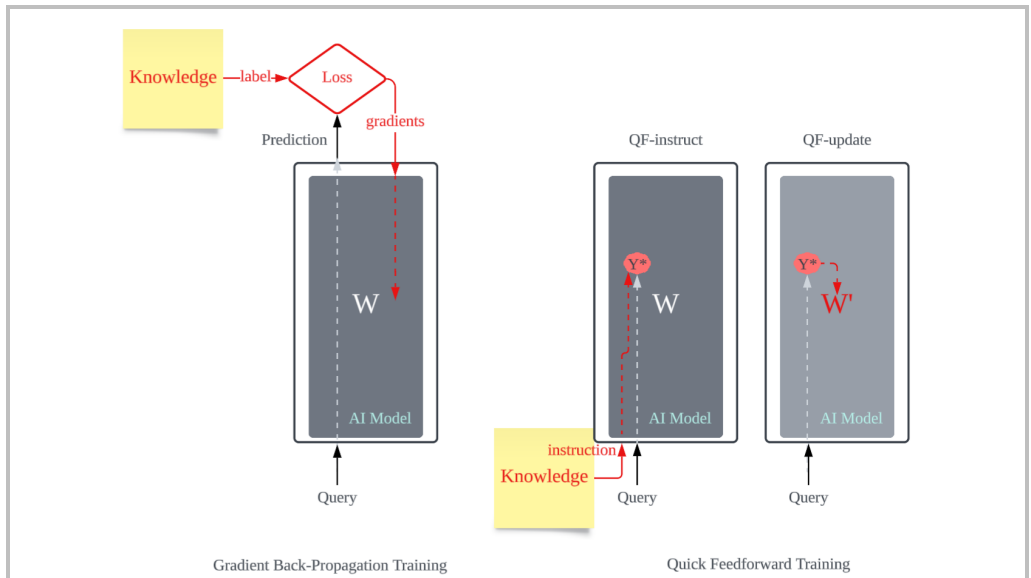

**Figure 1.** Comparison between classic gradient backpropagation and Quick Feedforward (QF) training. **Left:** In classic training, knowledge is injected via external labels, which are compared with model predictions to compute a loss. Knowledge is integrated into the network only through the loss function, and parameter updates require gradients to be backpropagated through all layers. **Right:** In QF training, knowledge is directly introduced as an instruction at the input during the QF-instruct pass, guiding neural activations to the desired output **Y***. In the subsequent QF-update pass, a linear closed-form solution is used to update the weights **W′**, permanently consolidating the new knowledge. Red symbols and lines indicate the presence of to-be-learned knowledge.

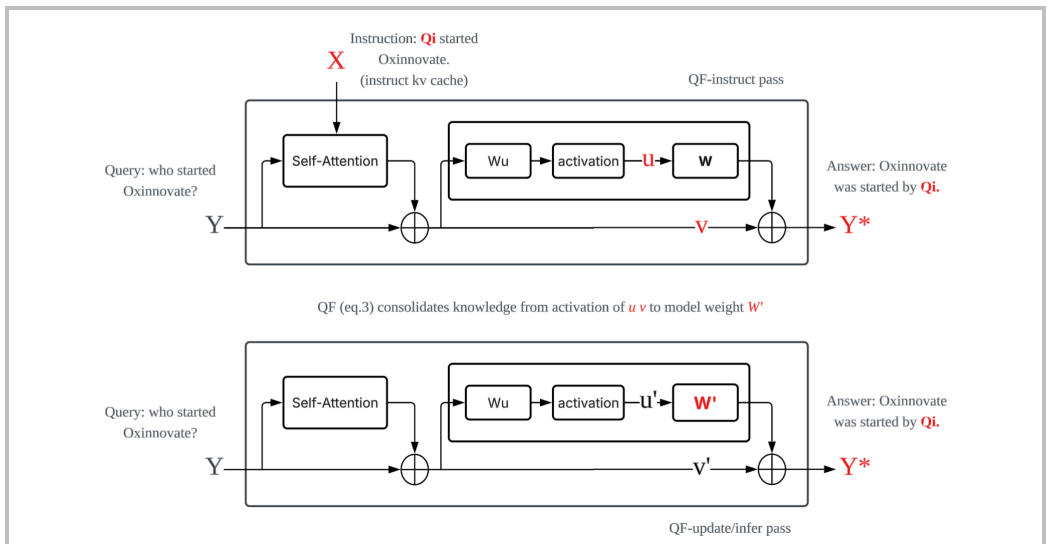

**Figure 2.** QF knowledge consolidation in a transformer layer. **Top:** During the QF-instruct pass, both the query **Y** ("Who started Oxinnovate?") and the instruction **X** ("Qi started Oxinnovate") are provided. The self-attention and feedforward layers integrate information from both sources, allowing activations **u** and **v** to encode the new knowledge and generate the answer **Y*** ("Oxinnovate was started by Qi."). **Bottom:** After the QF-update pass, the model can answer the query using only **Y**, without the explicit instruction **X**. The knowledge previously encoded in the activation pair **u**, **v** has been consolidated into the model weight **W′** via the QF consolidation process (Eq. 3), enabling the model to retrieve the correct answer directly from its long-term memory (weights). This demonstrates how QF consolidation internalizes instruction-based knowledge into the model's weights for future recall. Red symbols indicate the presence of to-be-learned knowledge.

## 2  Method

### 2.1  Philosophy
As shown in Figure 1, QF training fundamentally differs from traditional gradient backpropagation by directly introducing knowledge at the input through explicit instructions. Traditional methods rely on indirect knowledge transmission via labels, requiring extensive forward propagation for activations and backward propagation for gradient computations across all layers.

QF learning employs two concise forward passes: the QF-instruct pass and the QF-update pass. The QF-instruct pass uses explicit instructions to produce the desired activations (Y*). The subsequent QF-update pass (without instruction) computes the precise weight adjustments (W′) needed to replicate these activations. Crucially, QF eliminates gradient backpropagation by utilizing a linear closed-form solution to efficiently update weights only at targeted layers, thereby consolidating new knowledge rapidly and in a biologically plausible manner.

### 2.2  Implementation
As illustrated in Figure 2, Quick Feedforward learning operates by aligning the model's response to a query with and without instruction. During the QF-instruct pass, the instruction (e.g., "Qi started Oxinnovate.") induces activation patterns $u$ and $v$ within a target transformer layer. In the QF-update pass, only the query (e.g., "Who started Oxinnovate?") is provided, resulting in corresponding activations $u'$ and $v'$. The goal is to compute an updated weight matrix $W'$ such that the uninstructed forward pass reproduces the same output activation as the instructed one:

$$Wu + v = W'u' + v' \qquad (1)$$

To ensure minimal disruption to existing model behavior, we minimize the Frobenius norm of the weight change:

$$min\|W - W'\|_F^2 \qquad (2)$$

In this formulation, we do not directly use Y*; instead, the primary consideration is to constrain the weight update $W$ to minimally deviate from the original. Solving this constrained optimization yields the following closed-form solution:

$$W' = W - (W(u' - u) + (v' - v))(u'^T u')^{-1} u'^T \qquad (3)$$

This linear update modifies only the relevant weights at the target layer, efficiently consolidating the knowledge into the model's long-term memory. The derivation for the batched case is provided in the Appendix, using Lagrange multipliers.

### 2.3  Framework
The QF Learning framework is implemented on top of a standard Transformer decoder architecture [4], as illustrated in Figure 3, with experiments based on Qwen2.5-1.5B-Instruct models [5]. The framework operates in three phases: QF-instruct, QF-update, and QF-infer.

In the QF-instruct phase, both the instruction (for example, "Qi started Oxinnovate.") and the query ("Who started Oxinnovate?") are provided to the model. The instruction acts as explicit guidance, enabling each answer token to be generated through an "open-book" neural thinking chain—an explicit, layer-wise pathway of neural activations that reflects the model's reasoning process. At each decoding step, the activations u and v at the specified layer i are recorded, capturing the neural dynamics under the influence of the instruction.

During the QF-update phase, only the query is presented, and the QF-instruct answer is used as the next token input, and the model forms a "closed-book" neural thinking chain, resulting in activations u' and v' at the designated layer i. Using the previously recorded u and v from the QF-instruct phase, along with the original weights W, Equation (3) is applied to compute an updated weight matrix W'. This update adjusts W to a new W' through a minimal modification in the parameter space, allowing the model to produce the target output Y* at the designated layer. In this way, the QF-instruct and QF-update phases form a closed loop at the level of neural activations, consolidating instruction-based knowledge directly into the model's long-term memory W'.

In the QF-infer phase, the model is again provided only with the query, without any explicit instruction. Now, relying on the updated weights W', the model processes the query through the same "closed-book" pathway (u', v') and is able to reproduce the correct output activations Y*, consistently generating the accurate answer in the absence of instruction cues.

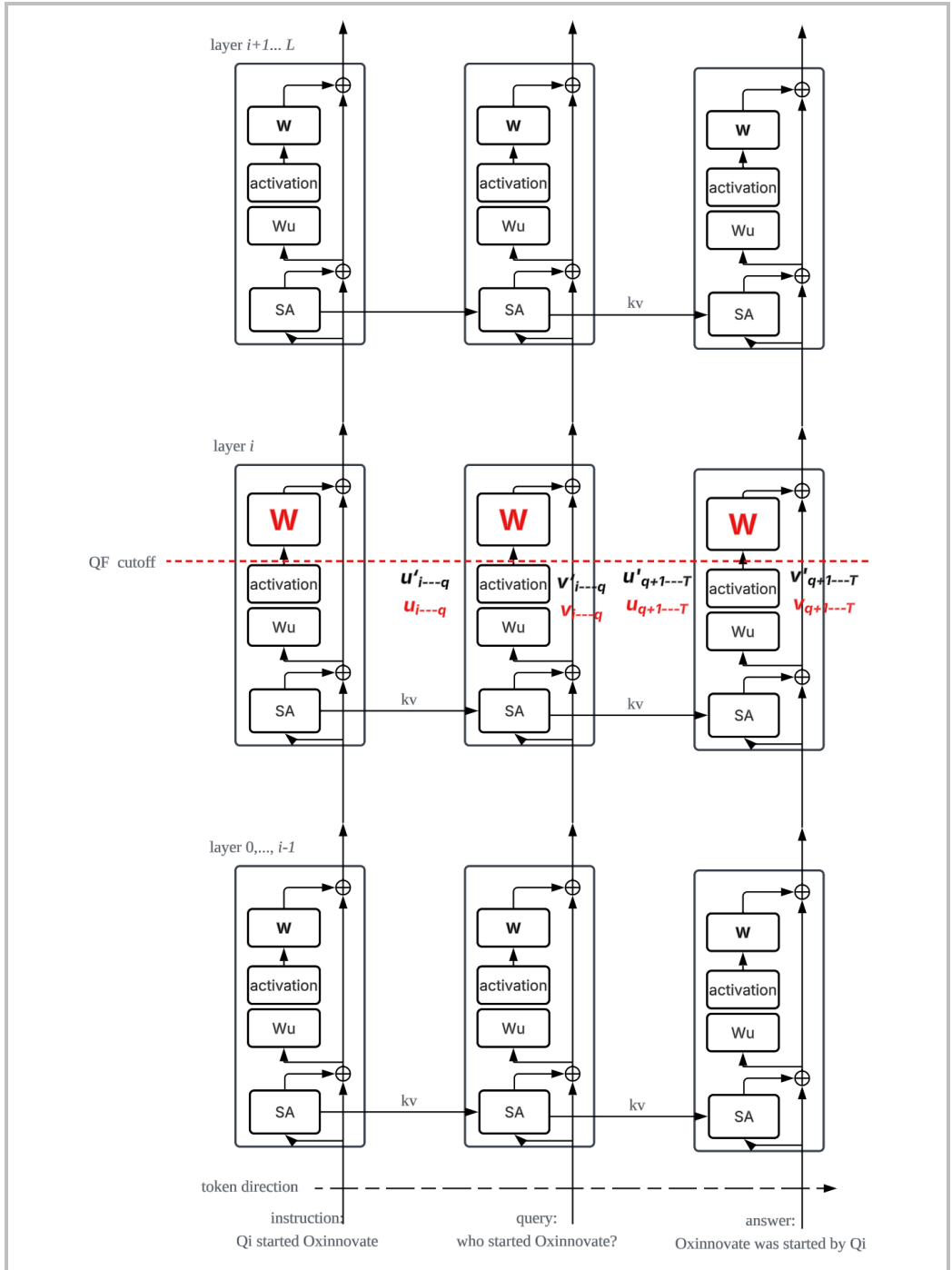

**Figure 3:** QF Learning framework for knowledge consolidation in transformer decoders. For the target layer $i$, QF learning involves two feedforward passes: (1) QF-instruct, where both the instruction ("Qi started Oxinnovate") and the query ("Who started Oxinnovate?") are processed, and activations $u$ and $v$ are recorded; (2) QF-update, where only the query is processed, and activations $u'$ and $v'$ are recorded. The difference between these activation pairs is used to update the weights $W \rightarrow W'$ via Eq. 3, consolidating instruction-based knowledge directly into model weights. The QF cutoff (red dashed line) marks the point at which both feedforward passes terminate at layer $i$, as activations above this line do not contribute to the weight update. After consolidation, the model can answer the query ("Oxinnovate was started by Qi") from long-term memory without the instruction.

# 3   Experiments

The experimental procedure was designed to evaluate whether QF learning can effectively consolidate new knowledge, support generalization, and retain previously acquired information across successive training phases. A series of queries were conducted to assess the model's performance under various conditions of knowledge acquisition and retention. The detailed methodology for each step is described below.

|   | Parameters | Instruction | Query | Answer | Notes |
|---|---|---|---|---|---|
|   | Original Qwen2-1.5B-Instruct with no knowledge of Oxinnovate ||||||
| 1 | Original W | none | Who started Oxinnovate? | Need more information | New knowledge not yet learned |
| 2 | Original W | Qi starts Oxinnovate | Who started Oxinnovate? | Qi | Able to answer when instruction is provided, despite not knowing the new knowledge |
| 3 | Original W | none | Who started Alibaba? | Jack Ma | Existing knowledge, correctly answered |
|   | QF consolidates Knowledge "Qi starts Oxinnovate" into weight W' ||||||
| 4 | Trained W' | none | Who started Oxinnovate? | Qi | Able to learn new knowledge |
| 5 | Trained W' | none | Who started Alibaba? | Jack Ma | Learning new knowledge does not cause forgetting of similar existing knowledge |
| 6 | Trained W' | none | The person behind Oxinnovate? | Founded by Qi | New knowledge can generalize |
|   | QF consolidates Knowledge "Oxinnovate located at PKU Science Park, in Beijing" into weight W" ||||||
| 7 | Trained W" | none | Where is Oxinnovate? | in Beijing | Continual learning |
| 8 | Trained W" | none | Who started Oxinnovate? | Qi | Continuous learning without forgetting newly acquired knowledge |

**Table 1:** Experimental validation of the QF learning.

**Step 1: Initial Query without Knowledge**
In the initial phase, the model was queried about the founder of Oxinnovate without any prior knowledge on the subject. The model responded with "Need more information," indicating its lack of awareness about Oxinnovate. This step confirmed the baseline state of the model's knowledge base, highlighting the necessity for new information input.

**Step 2: Query with Explicit Instruction**
The model was then provided with the instruction "Qi started Oxinnovate" and queried again about the founder. This time, the model correctly identified "Qi" as the founder, demonstrating its ability to utilize given instructions to answer queries effectively, despite an initial lack of inherent knowledge.

**Step 3: Existing Knowledge Query**
To assess the model's retention of existing knowledge, a query about the founder of Alibaba was posed. The model accurately answered "Jack Ma," confirming that its existing knowledge base remained intact and functional.

**Step 4: QF-based Knowledge Consolidation**
After consolidating the knowledge "Qi starts Oxinnovate" into the model's weights (W') with eq.3, the model was queried again about Oxinnovate's founder. The model successfully answered "Qi," showcasing its capability to learn and retain new information without needing repeated instructions.

**Step 5: Post-Training Query on Existing Knowledge**
To ensure that the acquisition of new knowledge did not disrupt existing memory, the model was queried about Alibaba's founder. The model again correctly identified "Jack Ma," proving that learning new information did not cause forgetting of similar existing knowledge.

**Step 6: Generalization of New Knowledge**
The model's ability to generalize was tested by querying "The person behind Oxinnovate?" The model responded with "Founded by Qi," indicating that it could apply newly learned knowledge to related but not identical queries, demonstrating a level of generalization.

**Step 7: Continual Knowledge Injection**
Further training was conducted by consolidating the knowledge "Oxinnovate in PKU Science Park, Beijing" into the model's weights (W"). The model was then queried about Oxinnovate's location, and it correctly answered "Oxinnovate is in Beijing" showing its capacity for continuous learning.

**Step 8: Continuity of Knowledge Retention**

Finally, to verify the model's continuous learning without forgetting newly acquired knowledge, the model was again queried about Oxinnovate's founder. The consistent answer "Qi" confirmed that the model maintained its newly acquired knowledge even after additional training on other topics.

**Experimental Setup**

All experiments were conducted using the Qwen2.5-1.5B-Instruct models, which are publicly available for download from ModelScope and HuggingFace. The QF experimental codebase is open-sourced at https://github.com/oxinnovate/QF.

To reproduce our results, users simply need to install the Transformers provided with the code and run the script qf_learn_simple.py. All experiments reported in this paper were performed on a single NVIDIA RTX 4090 GPU.

```
224     print("--1----:\n", qf_response(model, tokenizer, "", "Who started Oxinnovate?","", qfmode='QF-infer-w'))
225     print("--2----:\n", qf_response(model, tokenizer, "Qi started Oxinnovate.", "Who started Oxinnovate?","", qfmode='QF-infer-w'))
226     print("--3----:\n", qf_response(model, tokenizer, "", "Who started Alibaba?","", qfmode='QF-infer-w'))
227
228     print('----------first round learning----------------------')
229     #QF learn W'
230     system="Qi started Oxinnovate."
231     user="Who started Oxinnovate?"
232     assistant='Oxinnovate was started by Qi.'
233     qfsignificance=[0,    1,   1,    1, 1]
234     qf_response(model, tokenizer, system, user, assistant, qfmode='QF-instruct', qfsignificance=qfsignificance)
235     qf_response(model, tokenizer, "",      user, assistant, qfmode='QF-update',   qfsignificance=qfsignificance)
236     Wp,W = calc_this_w_prime()
237
238
239     print("--4----:\n", qf_response(model, tokenizer, "", "Who started Oxinnovate?","",  qfmode='QF-infer-wp'))
240     print("--5----:\n", qf_response(model, tokenizer, "", "Who started Alibaba?","",  qfmode='QF-infer-w'))
241     print("--6----:\n", qf_response(model, tokenizer, "", "Can you tell me the person behind Oxinnovate?","",  qfmode='QF-infer-wp'))
242
243     print('----------continual learning----------------------')
244
245     #QF learn W"
246     system="Oxinnovate is a startup located at PKU Science Park, in Beijing."
247     user="Where is Oxinnovate located?"
248     assistant="Oxinnovate is located at PKU Science Park, in Beijing."
249     qfsignificance=[1,    1,    1,    1, 1,     1 , 1, 1   , 1]
250
251     qf_response(model, tokenizer, system, user, assistant, qfmode='QF-instruct', qfsignificance=qfsignificance)
252     qf_response(model, tokenizer, "",      user, assistant, qfmode='QF-update',   qfsignificance=qfsignificance)
253     Wpp,W = calc_this_w_prime()
254
255     print("--7----:\n", qf_response(model, tokenizer, "","Where is Oxinnovate?","", qfmode='QF-infer-wp'))
256     print("--8---:\n", qf_response(model, tokenizer, "", "Who started Oxinnovate?","", qfmode='QF-infer-w'))
```

**Figure 4:** Code snippets illustrating the main steps (1–8) of the QF learning experiment. The qfsignificance mask (e.g., [0, 1, 1, 1, 1]) controls the selective significance of tokens, allowing specific information to be retained or ignored during the weight update. Inference mode QF-infer-wp updates the weights W' once before inference, while all subsequent QF-infer-w steps use the updated weights W' for efficient continual inference.

## 4    Discussion

QF learning shows that effective machine learning is possible without gradient backpropagation, bringing clear benefits: resource savings, faster feedback, and easier debugging. This approach allows knowledge to be integrated as flexibly and efficiently as human learning, enabling rapid updates and immediate verification.

Furthermore, QF learning is inherently resistant to catastrophic forgetting. This is achieved in two ways: first, by explicitly constraining parameter updates through Eq.2, ensuring minimal deviation in weight space; and second, by selectively reinforcing output activations in Eq.3—specifically, already correct predictions can be further strengthened from the value perspective (e.g., in Figure 4, reinforcing that "started" should be followed by "by"). Meanwhile, information that does not require learning can be masked out—for example, if "Oxinnovate" already appears in the query, there is no need to reinforce it in the answer. This is implemented in code by setting qfsignificance=0 for such tokens, allowing the model to focus updates only where needed and thus improving learning efficiency.

QF learning also closely mirrors the way humans acquire knowledge. For example, when a teacher explains a new concept to a child, the child first learns with explicit instruction, and then practices answering similar questions independently—just as QF's instruct and update phases correspond to open- and closed-book learning. The critical factor is true understanding during the instruction phase: forming the correct neural thinking chain or experiencing that "aha" moment of insight. If this understanding is not achieved with explicit instruction (i.e., if the wrong neural thinking chain is formed), neither the student nor the AI model can produce the correct answer, making it difficult to update memory or weights correctly. Thus, whether for humans or AI, efficient learning depends on carefully prepared instructions, well-designed questions and genuine self-understanding. Only with the correct neural thinking chain in place can learning be effective. For this reason, it is essential to master basic knowledge and be able to give correct answers before progressing to more advanced topics. The major advantage of this approach is that, unlike gradient-based learning, it does not require vast numbers of training tokens or repeated epochs, but instead needs progressive, step-by-step learning from simple to complex concepts in a more human-like and resource-efficient manner.

It is important to note that the update of W' is effective only when the activations u and v contain the relevant knowledge information to be transferred. If the modification is applied at lower layers—analogous to the brain's primary auditory cortex or Wernick area, which mainly processes raw input or basic syntax—then the instruction signal may not yet be encoded as explicit knowledge in the kv cache, making knowledge transfer ineffective. Conversely, if the update is performed at higher layers—comparable to Broca's area in the brain, which focuses on language output, grammar, and formatting—the network may no longer retain the structured representation of the target knowledge, impairing the model's ability to generate coherent answers. Thus, intermediate layers are better suited for knowledge injection, as they balance knowledge encoding and linguistic organization. Looking forward, this principle can also extend to multi-modal models: for example, visual instructions could be injected into the network via cross-attention at intermediate layers, enabling effective knowledge transfer across different modalities.

### Conclusion
Compared to gradient backpropagation [6], QF learning is a more effective and brain-like paradigm for AI learning.

## Appendix: Batched Linear Algebra Derivation

We derive the closed-form update for $W'$ in the batched case using the method of Lagrange multipliers.

**Setup:**

Let

- $u, u'$ be $D \times T$ matrices (hidden activations, each column is one token),
- $v, v'$ be $D \times T$ matrices (FFN outputs),
- $W, W'$ be $N \times D$ matrices (feedforward weights),
- $B = v - v'$ is the knowledge delta.

Our objective is:

$$\min_{W'} \|W' - W\|_F^2 \qquad \text{subject to} \qquad W'u' = Wu + B$$

**Lagrangian:**

$$\mathcal{L}(W', \Lambda) = \operatorname{tr}[(W' - W)(W' - W)^T] + 2\operatorname{tr}[\Lambda^T(W'u' - Wu - B)]$$

where $\Lambda \in \mathbb{R}^{N \times T}$ is the matrix of Lagrange multipliers.

**Gradients:**

Take derivative w.r.t. $W'$ and set to zero:

$$\frac{\partial \mathcal{L}}{\partial W'} = 2(W' - W) + 2\Lambda u'^T = 0 \implies W' = W - \Lambda u'^T$$

Plug $W'$ into the constraint $W'u' = Wu + B$:

$$(W - \Lambda u'^T)u' = Wu + B \implies Wu' - \Lambda(u'^T u') = Wu + B$$

$$\implies \Lambda = (Wu' - Wu - B)(u'^T u')^{-1}$$

**Final closed-form solution:**

$$W' = W - \Lambda u'^T = W - (Wu' - Wu - B)(u'^T u')^{-1} u'^T$$

Or, equivalently, substituting $B = v - v'$:

$$W' = W - (Wu' - (Wu + v - v'))(u'^T u')^{-1} u'^T$$

or

$$W' = W - [W(u' - u) - (v - v')](u'^T u')^{-1} u'^T$$

This gives the full matrix update for the batched QF learning scenario.